%% file: AECR.tex
\documentclass[sigconf, review, nonacm]{acmart}
\usepackage{pvldb}

\renewcommand\vldbdoi{XX.XX/XXX.XX}
\renewcommand\vldbpages{XXX-XXX}
\renewcommand\vldbavailabilityurl{https://anonymous.4open.science/r/AECR-AFAF}

\usepackage{subcaption}
\usepackage{subfiles}
\usepackage{fontawesome}
\usepackage{multirow}
\usepackage{booktabs}
\usepackage{tabularx} 
\usepackage{makecell} 
\usepackage{url}
\usepackage{svg} 

\usepackage{enumitem,cleveref,amsmath}
\usepackage{soul}

\usepackage[most]{tcolorbox}

\theoremstyle{definition}

\usepackage{xcolor}    
\usepackage{colortbl}  
\usepackage{multirow}  

\renewcommand{\arraystretch}{1}
\AtBeginDocument{%
  }%

\begin{document}

\title{Abstract Event Causal Rules: Induction and Application}

\author{Ziwei Zheng}
\orcid{0009-0000-1348-7338}
\affiliation{%
    \institution{School of Electronic Information and Communications, Huazhong University of Science and Technology}
    \city{Wuhan}
    \country{China}
}
\email{ziweizheng@hust.edu.cn}
\author{Peiqiong Chen}
\orcid{0009-0003-4218-1161}
	\affiliation{%
		\institution{School of Electronic Information and Communications, Huazhong University of Science and Technology}
		\city{Wuhan}
		\country{China}
	}%
\email{pqchen@hust.edu.cn}

\author{Bang Wang}
\orcid{0000-0002-0312-4805}
	\affiliation{%
		\institution{School of Electronic Information and Communications, Huazhong University of Science and Technology}
		\city{Wuhan}
		\country{China}
	}%
\email{wangbang@hust.edu.cn}

\begin{abstract}
Event-centric intelligent analytical systems heavily depend on explicit causal event knowledge for risk early warning, decision-making support and narrative comprehension. Nevertheless, existing instance-level causal pairs suffer severe generalization deficits on low-frequency long-tail and unseen event combinations. To address this limitation, this work proposes Abstract Event Causal Rule (AECR), a novel relation-level causal abstraction paradigm that transforms concrete cause-effect pairs into generalized abstract causal logic while retaining their intrinsic causal relationships. We design a multi-agent Concrete-to-Abstract Causal Induction (CACI) system coupled with similarity-constrained clustering to distill trustworthy AECRs from noisy raw causal data, based on which two complete AECR knowledge bases are built. To validate the practical utility of abstract causal knowledge, we propose an Abstract Rule-Guided Causal Attention Encoder (AR-GCAE), which injects the retrieved AECRs into the causality Graph Event Prediction (CGEP) benchmark task via rule-guided attention layers and gated representation fusion. Quantitative experimental results reveal that applying AECRs substantially strengthens the generalization capacity of event causal reasoning and brings consistent performance improvements to event prediction, with the most prominent gains observed on rare and unseen event samples.

\end{abstract}


\maketitle

\vldbtopmatter

\input{Sections/Introduction}

\input{Sections/RelatedWork}

\input{Sections/Construction}

\input{Sections/ConstructionEvaluation}

\input{Sections/Application}

\input{Sections/ApplicationExperiment}

\input{Sections/Conclusion}

\bibliographystyle{ACM-Reference-Format}
\bibliography{reference}

\end{document}

%% file: Sections/Introduction.tex
\section{Introduction}\label{Sec:Introduction}

Future event prediction relies not merely on the memorization of historical event observations, but crucially on causal knowledge that characterizes the causal dependencies between distinct events~\cite{lv-etal-2020-integrating}. For example, the causal knowledge \textsf{'natural disaster'} \(\rightarrow\) \textsf{'property damage'} lets a system foresee damage from a rarely seen disaster like \textsf{'freezing rain'}, even without any prior instance of it. Equipping intelligent systems with generalizable event causal knowledge has long been a core objective of event-centric artificial intelligence, which underpins a broad spectrum of high-impact applications including risk early warning, intelligent decision support, and automated narrative understanding~\cite{WhatisEKG, liu2020event}. This raises a fundamental research question: \textit{what standardized form of causal knowledge should machines learn and adopt to achieve robust generalization across open-world diverse event scenarios?}

\par
Existing event causal knowledge is predominantly concrete and instance-level in current research paradigms. Whether extracted as causal pairs or organized as \textit{event causality graphs} (ECGs)~\cite{zhan2024would}, each causal fact is tightly bound to specific storylines~\cite{cao2021knowledge}, with causes and effects articulated via context-specific event descriptions. Such instance-level causal knowledge exhibits inherent limitations in transferability: causal facts derived from one specific scenario rarely generalize to semantically similar yet lexically divergent event cases. Accordingly, models built upon instance-level causal knowledge tend to memorize superficial lexical patterns and statistical co-occurrences. They maintain reasonable performance on frequent, observed event instances but struggle severely with long-tail and unseen events that prevail in practical real-world applications~\cite{du2022resin, lyu2021zero}. This fundamental deficiency arises because existing methods lack abstract, transferable causal mechanisms underlying diverse concrete instances, preventing robust causal generalization beyond observed individual scenarios.

\par
A natural mitigation to the above limitation lies in \textit{event abstraction}, which maps concrete event mentions to high-level conceptual representations to facilitate knowledge sharing across superficially distinct instances. Popular commonsense knowledge resources including ATOMIC, GLUCOSE, and ACCESS~\cite{mostafazadeh2020glucose, sap2019atomic, vo2025access} have validated that abstract conceptual knowledge can effectively improve cross-instance generalization. Nevertheless, existing resources conduct event abstraction in \textit{isolation}: each event mention is converted into a generic concept independently, without accounting for its paired causal counterpart. Such isolated single-event abstraction inevitably introduces \textit{semantic drift}. Once an event is abstracted into an overly general concept, it loses the critical semantic facets that maintain the original causal dependency, rendering the abstracted cause and effect unable to form a logically valid causal relation. As an example, \textsf{'heavy rainfall'} \(\rightarrow\) \textsf{'flash flood'} is a reliable causal relation, yet isolatedly abstracting the cause into \textsf{'weather condition'} disregards whether this causal linkage is preserved, thereby breaking the causal dependency: a generic \textsf{'weather condition'} no longer entails a \textsf{'flash flood'}.

\par
To overcome the critical limitations of isolated single-event abstraction, we argue that causal abstraction should take the complete \textit{causal relation} as the basic processing unit. Specifically, the abstraction process must jointly encode paired cause and effect events to align their conceptual granularities, while fully retaining the intrinsic causal linkage between them. Following this core principle, we formalize a novel abstract paradigm named \textbf{Abstract Event Causal Rule} (AECR). Each AECR denotes a transferable causal pattern (e.g., \textsf{'natural disaster'} $\rightarrow$ \textsf{'property damage'}) distilled from abundant concrete event pairs. It can serve as reusable prior causal knowledge and support robust knowledge transfer across event pairs with distinct lexical expressions yet identical causal logic. The central objective of this work is on how to induce abstract rules to build high-quality AECR knowledge bases and verify that such rules deliver dependable, practically viable causal knowledge to facilitate event causal reasoning.

\par
However, constructing a high-quality AECR knowledge base from noisy instance-level causal pairs poses substantial challenges. Directly prompting a Large Language Model (LLM) to abstract raw causal pairs in a single pass easily leads to hallucinations and co-occurrence bias~\cite{zevcevic2023causal, li2023open, liang2024encouraging, shi2024replug}. To address this challenge, we design a multi-agent \textit{Concrete-to-Abstract Causal Induction} (CACI) system. It consists of five dedicated functional agents, namely restatement, causal gatekeeping, abstraction, selection and judgment agents, which collaborate within an iterative feedback loop to derive trustworthy relation-level causal logic for concrete cause-effect pairs. Afterward, all derived causal logic are grouped via agglomerative hierarchical clustering with intra-cluster similarity constraints. Each cluster is further distilled by an LLM to generate one unified abstract causal rule. We apply the proposed CACI system to two standard ECG benchmark datasets to build two AECR knowledge bases. Human evaluation results demonstrate that the constructed AECR bases achieve high rationality, strong discriminability, and practical usability. To make these rules readily reusable, we further train an AECR retriever for each knowledge base that, given a concrete cause-effect pair, retrieves its most explanatory abstract rules from the knowledge base. It can serve as a plug-and-play module to inject transferable causal priors into downstream reasoning.

\par
Beyond evaluating the intrinsic quality of the induced abstract rules, the practical value of AECR knowledge bases ultimately hinges on their performance for downstream causal reasoning tasks. To empirically verify such practical applicability, we employ the \textit{Causality Graph Event Prediction} (CGEP) task~\cite{zhan2024would} as a representative evaluation benchmark, which aims to predict the subsequent event triggered by a given anchor event within an ECG structure. We note that the CGEP task merely serves as a diagnostic probe to quantify the utility of AECR knowledge: our core objective is to examine whether the incorporation of AECR knowledge can boost model causal reasoning capability. We devise an Abstract Rule-Guided Causal Attention Encoder (AR-GCAE) for event prediction: it first encodes ECG structures via a topology-aware Transformer; it then leverages the latent representation of the unobserved target event to retrieve relevant abstract rules. Afterwards, the retrieved rules are dynamically injected into graph embeddings via a rule-guided attention layer, followed by gated fusion to balance rule-agnostic and rule-enhanced representations. 
Experiments on two public datasets are conducted and consistent performance gains over the state-of-the-art methods are observed. Notably, the most substantial performance improvements appear on rare and unseen events. Experimental outcomes offer direct empirical proof that AECR can provide transferable causal knowledge unavailable from raw concrete event pairs alone.

\par
In summary, this work delivers four core contributions:
\begin{itemize}[leftmargin=*]
    \item We formalize the Abstract Event Causal Rule (AECR), a relation-level abstraction that extracts transferable causal mechanisms from concrete event pairs, and construct two high-quality AECR knowledge bases as reusable causal priors.
    \item We design the Concrete-to-Abstract Causal Induction (CACI) system to extract credible abstract causal rules from noisy event instances, whose rationality, discriminability and utility are verified via human evaluation.
    \item We devise the Abstract Rule-Guided Causal Attention Encoder (AR-GCAE), which integrates retrieved AECRs into the CGEP task through rule-aware attention and gated fusion to verify the practical value of AECR knowledge base.
    \item We conduct experiments against state-of-the-art methods and find our AR-GCAE achieves steady performance gains, particularly on rare and unseen events, which confirms the superior transferability of AECR over raw concrete event data.
\end{itemize}


    

%% file: Sections/RelatedWork.tex
\section{Related Work}\label{Sec:RelatedWork}


\subsection{Commonsense and Event Knowledge}
A long line of resources injects generalizable priors into event reasoning by encoding commonsense and causal knowledge. Entity-centric graphs, such as the ConceptNet~\cite{speer2017conceptnet}, organize taxonomic relations among~\cite{jiayang2024eventground} concepts, and event-centric resources, such as the ATOMIC~\cite{sap2019atomic} and GLUCOSE~\cite{mostafazadeh2020glucose}, record if-then inferential and generalized causal rules around everyday events; however, they abstract each event in isolation, grouping neither mentions by causal equivalence nor relations into coherent causal structure, so an event lifted to an overly generic concept loses the specific facet that should have sustained its original causal link, causing a semantic drift under which the abstracted cause and effect no longer compose into a valid causal statement. A second line instead builds causal graphs over abstract events: ACCESS~\cite{vo2025access} lifts daily-life events to an abstraction level and connects hundreds of event abstractions into a commonsense causal graph, whereas text-derived resources, such as the CauseNet~\cite{heindorf2020causenet} and CRAB~\cite{romanou2023crab}, mine causal pairs directly from documents; yet the former abstracts events node-by-node for causal discovery rather than modeling the causal relation itself, and the latter provides no abstraction at all, leaving fine-grained graphs that explode in size and stay bound to specific mentions. What is still missing, therefore, is abstraction anchored on the causal relation itself, jointly elevating cause and effect while preserving the causal force between them.

\subsection{Causality Graph Event Prediction}
Event prediction has progressed from reasoning over linear event chains toward reasoning over structured event graphs~\cite{li2018constructing}. Early script event prediction formulations~\cite{chambers2008unsupervised, granroth2016happens, pichotta2016learning} represent history as an ordered narrative chain and forecast the next event, first via statistical co-occurrence counts and later via dense and pre-trained language-model encoders; a single chain, however, cannot express the branching, many-to-many causal dependencies of real narratives, and sentence-level encoding further constrains discourse-level reasoning. To expose richer structure, historical events and their causal relations are first extracted from documents into an ECG~\cite{tao2023seag}, over which the CGEP task~\cite{zhan2024would} forecasts the consequential event of an anchor event. Methods along this line broadly divide into graph-based encoders~\cite{zhan2025, ding2019event, jiayang2024eventground} that propagate evidence along causal edges but initialize nodes with static embeddings underexploiting event semantics, and language-model-based encoders~\cite{zhan2024would, jiang2023structgpt, wu2025graph} that linearize the graph into prompts to capture context yet thereby distort the intrinsic graph topology; more recent efforts further augment the graph with LLM-generated nodes and edges and apply robust training to counter its structural deficiency~\cite{zheng2026, luo2024reasoning}, though the knowledge they exploit still stays bound to concrete training-graph instances. How to equip event prediction with causal knowledge that generalizes beyond such concrete instances to low-frequency and unseen events still remains an open problem.

%% file: Sections/Construction.tex
\section{Induction of Abstract Event Causal Rules}
\label{Sec:Construction}
Let $\mathcal{P} = \{p_1, p_2, \dots, p_N\}$ denote the set of causal event pairs extracted from a set of event causality graphs. Each pair $p_i = (e_i^\texttt{c}, e_i^\texttt{e})$ represents a directed causal relation where the cause event $e_i^\texttt{c}$ leads to the effect event $e_i^\texttt{e}$. Each event $e_i$ is defined as a tuple $e_i = (m_i, s_i)$, comprising an event mention $m_i$ (a word or phrase) and the sentence $s_i$ containing the mention to provide contextual background. The objective is to induce an AECR knowledge base $\mathcal{R}$ from $\mathcal{P}$.

\par
We design a \textit{Concrete-to-Abstract Causal Induction} system (CACI) to build an AECR knowledge base from event causality graphs, consisting of two steps: (1) Extract concrete causal logics at the instance level from all causal event pairs; (2) Generate abstract causal rules from clustered concrete instance-level causal logics. Furthermore, we design an AECR retriever based on the constructed AECR knowledge base for downstream applications.


\begin{figure*}[htbp]  
	\centering
	\includegraphics[width=\textwidth]{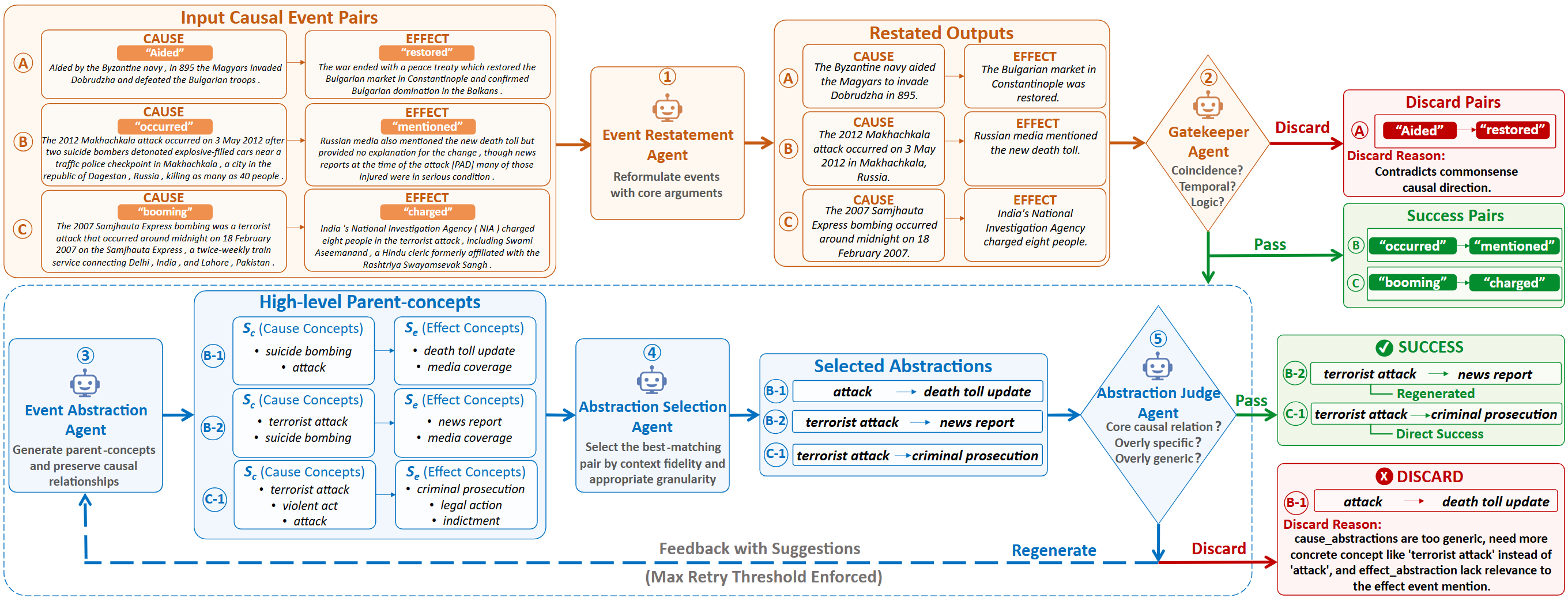}
	\caption{Overview of the CACI multi-agent system for concrete event causal logic extraction. The five agents run sequentially with an iterative feedback loop, illustrated by three representative traces: a discarded pair (Example-A), acceptance after feedback-driven refinement (Example-B), and direct success (Example-C).}
	\label{fig:mas_pipeline}  
\end{figure*}
\subsection{Concrete Event Causal Logic Extraction}
We first extract the concrete causal logic at the instance level from each pair of causal events. Figure~\ref{fig:mas_pipeline} presents the overall pipeline of our LLM-driven multi-agent system, which integrates both the sequential agent workflow and representative examples of concrete event causal logic extraction, illustrating typical execution traces including discarded cases, iterative refinement with feedback, and direct success cases. In particular, the system employs a feedback loop that iteratively refines the intermediate abstractions until a valid causal logic emerges. The pipeline consists of the following agents, which are executed sequentially. We note that the operation of each agent is executed as prompting an LLM for the output~\footnote{Due to the page length limit, all prompts are detailed in the supplementary material.}.

\par
$\bullet$ \textbf{Event Restatement Agent.} This agent reformulates each event into a concise statement with event-focused arguments, yet removing redundant descriptions. Particularly, given an event $e_i = (m_i, s_i)$ with event mention  $m_i$ and context sentence $s_i$, the agent first extracts the event-focused arguments, like participants, time, location, etc.,  and then generates an event-focused restatement while discarding other event-irrelevant details. For example, from the context sentence of the effect event of the \textsf{Example-A} in Figure~\ref{fig:mas_pipeline}, the agent generates a restatement that retains the core argument of \textsf{'restored'} with regard to the \textsf{'Bulgarian market in Constantinople'}, while removing other event-irrelevant  narratives such as the \textsf{'war'} and the \textsf{'peace treaty'}. The resulting restatement is more complete than the raw event mention, yet more event-focused than the original context sentence for downstream causal reasoning.

\par
$\bullet$ \textbf{Gatekeeper Agent.} This agent performs a causal plausibility screening on each causal event pair. Specifically, the agent takes as input the event mention, its restated form, and the original context for both cause event and effect event, and evaluates the causal plausibility of the event pair against three criteria: (1) the cause must temporally precede the effect, (2) the causal link should not be a mere coincidence, and (3) the effect must be a logically expected consequence of the cause. In the \textsf{Example-A}, the cause event describes a military invasion aided by external forces, while the effect event describes the restoration of a trade market in the invaded region. Since invasions typically lead to disruption rather than restoration, the effect contradicts the expected causal direction. The agent therefore issues a \textsf{'DISCARD'} decision, preventing this invalid pair from entering the downstream abstraction agent.

\par
$\bullet$ \textbf{Event Abstraction Agent.} This agent abstracts concrete events into high-level parent-concepts while preserving causal relationships. Specifically, taking the event mention, its restated form, and the original context of both cause event and effect event as input, the agent produces multiple parent-concepts for each event to capture distinct facets of the event. For instance, in the \textsf{Example-B}, the agent abstracts the cause event into \textsf{'terrorist attack'} from a security perspective and \textsf{'suicide bombing'} from a tactical perspective, yielding $\mathcal{S}_\text{c}$ = $\{\textsf{'terrorist attack'}$, $\textsf{'suicide bombing'}\}$, and abstracts the effect event accordingly, yielding $\mathcal{S}_\text{e}$ = $\{\textsf{'news report'}$, $\textsf{'media coverage'}\}$. The candidate sets $\mathcal{S}_\text{c}$ and $\mathcal{S}_\text{e}$  contain the abstracted parent-concepts for the cause event and effect event, respectively, which are then passed to downstream agents for selection and validation. Moreover, the agent can regenerate abstractions iteratively by incorporating feedback from downstream agents. To avoid excessive iterations, a maximum retry threshold is enforced.

\par
$\bullet$ \textbf{Abstraction Selection Agent.} This agent, for a given event pair, selects the best-matching pair of parent-concept abstractions from the candidate sets $\mathcal{S}_c$ and $\mathcal{S}_e$. It searches over all combinations of a cause abstraction $\tilde{c} \in \mathcal{S}_c$ and an effect abstraction $\tilde{e} \in \mathcal{S}_e$, and identifies the pair that best captures the core causal force, guided by two criteria: (1) Context fidelity, i.e., alignment with the original event descriptions; and (2) Appropriate granularity, i.e., the abstraction should be neither too specific to merely restate the original event nor too generic to preserve causal link between the abstracted cause and effect. In the \textsf{Example-B}, from $\mathcal{S}_c$ = $\{\textsf{'terrorist attack'}$, $\textsf{'suicide bombing'}\}$ and $\mathcal{S}_e$ = $\{\textsf{'news report'},$ $ \textsf{'media coverage'}\}$, the agent selects $\textsf{'terrorist attack'}$ $\rightarrow$ $\textsf{'news report'}$. This choice respects both criteria: \textsf{'terrorist attack'} is preferred over \textsf{'suicide bombing'} for its more appropriate granularity, as the latter merely restates the concrete tactic; likewise, \textsf{'news report'} is favored over \textsf{'media coverage'} for its closer fidelity to the reporting act described in the original context. The selected pair is then passed to the next judge agent for validation.

\par
$\bullet$ \textbf{Abstraction Judge Agent.} This agent judges whether the selected pair of abstract parent-concepts conforms to sound causal logic. Specifically, it examines the selected pair of parent-concept abstractions $\tilde{c}_i^{\ast}$ $\rightarrow$ $\tilde{e}_i^{\ast}$ against three criteria. First, the abstraction pair should neither be overly specific, i.e., merely a synonym of the original event, nor overly generic, i.e., too broad to retain causal meaning. Second, each abstraction should reflect its corresponding event, i.e., it must not distort from the semantics of the original event. Third, the selected abstraction pair should embody the core causal link of the original event pair, enabling deduction of the abstract effect from the abstract cause. If all criteria are satisfied, the agent issues a \textsf{'SUCCESS'} decision. Otherwise, it issues a \textsf{'DISCARD'} decision with feedback for the Event Abstraction Agent to regenerate parent-concept abstractions, and the process of Abstraction, Selection and Judge iterates till the maximum tries.

\par
Through this iterative feedback loop, an abstraction may be accepted on the first try, accepted after rounds of refinement, or ultimately discarded if it still fails once the maximum number of retries is reached. As an instance of direct acceptance, in the \textsf{Example-C} the selected pair \textsf{'terrorist attack'} $\rightarrow$  \textsf{'criminal prosecution'} passes the judge agent without any refinement. The \textsf{Example-B} illustrates acceptance after a single round of refinement. In the first round, the Event Abstraction Agent produces $\mathcal{S}_\text{c}$ = $\{\textsf{'suicide bombing'}$, $\textsf{'attack'}\}$ and $\mathcal{S}_\text{e}$ = $\{\textsf{'death toll update'}$, $\textsf{'media coverage'}\}$, and the Abstraction Selection Agent selects \textsf{'attack'} $\rightarrow$ \textsf{'death toll update'}. The judge agent rejects this pair on two grounds: the cause abstraction \textsf{'attack'} is too generic and should be a more concrete concept such as \textsf{'terrorist attack'}; and the effect abstraction \textsf{'death toll update'} inaccurately reflects the original event, whose effect is that the media \textsf{'mentioned'} the toll. Guided by this feedback, in the second round the regenerated candidate sets change to $\mathcal{S}_\text{c}= \{\textsf{'terrorist attack'}, $ $ \textsf{'suicide bombing'}\}$ and $\mathcal{S}_\text{e}=\{\textsf{'news report'}$ $, $ $ \textsf{'media coverage'}\}$, and the newly selected pair $\textsf{'terrorist attack'} \rightarrow \textsf{'news report'}$ passes the judge agent in the second round.

\par
$\bullet$ \textbf{Output.} The aforementioned pipeline outputs a set of concrete event causal logics, denoted as $\mathcal{A} = \{a_1, \dots, a_M\}$, where $a_j = (\tilde{c}_j^{\ast}, \tilde{e}_j^{\ast})$ is called an instance-level concrete event logic for the causal event pair $p_j= (e_j^\texttt{c}, e_j^\texttt{e})$ in datasets.


\subsection{Abstract Event Causal Rule Generation}
\label{Subsec:AECRGeneration}
After extracting $M$ concrete causal logics from $N$ causal event pairs, we first apply a clustering approach to group these logics into clusters and then leverage an LLM to generate an abstract event causal rule for each cluster.

\par
We leverage a frozen text encoder (e.g., RoBERTa~\cite{liu2019roberta}) to first encode each concrete causal logic $a_j = (\tilde{c}_j^{\ast}, \tilde{e}_j^{\ast})$ into a vector representation $\mathbf{v}_j$ as follows. The input textual template $t_j$ for the text encoder is constructed by
\[
    t_j = \texttt{[CLS]} \oplus \tilde{c}_j^{\ast} \oplus \text{``causes''} \oplus \tilde{e}_j^{\ast} \oplus \texttt{[SEP]}
\]
where $\oplus$ denotes sequence concatenation, and $\texttt{[CLS]}$ and $\texttt{[SEP]}$ are special tokens indicating the start and end of the sentence. Next, we apply a mean pooling operation exclusively over the encoder output embeddings of the non-special tokens to obtain $\mathbf{v}_j$ as follows:
\[
    \mathbf{v}_j = \frac{1}{|\mathcal{T}_j|} \sum_{k \in \mathcal{T}_j} \mathbf{h}_{j,k},
\]
where $\mathbf{h}_{j,k}$ represents the encoded hidden state of the $k$-th token, and $\mathcal{T}_j$ denotes the index set of all non-special tokens within the input template $t_j$.

\par
Using these logics' representations $\{\mathbf{v}_j\}$, we calculate the pairwise similarity between any two causal logics $a_u$ and $a_v$ to construct a distance matrix $\mathbf{D}$ for the subsequent clustering process:

\[
    \mathbf{D}_{u,v} = 1 - \frac{\mathbf{v}_u \cdot \mathbf{v}_v}{\|\mathbf{v}_u\| \|\mathbf{v}_v\|}
\]
Based on $\mathbf{D}$, we employ the agglomerative hierarchical clustering with the complete-linkage criterion to group these concrete causal logics into clusters. During this process, we enforce an \textit{intra-cluster similarity threshold} $\tau$ to ensure robust cohesion. Finally, we filter out clusters containing fewer than $\mu$ members, as these are considered to represent weakly generalizable patterns that are unlikely to yield reliable abstract rules.


We then leverage an LLM (e.g., Gemini-3.1-Pro) to generate an abstract event causal rule for each cluster. Given that the concrete causal logics within a cluster share a common causal pattern, the LLM is prompted to distill this shared logic into a generalizable rule. The prompt is designed with three key principles: capture the dominant causal pattern shared by the majority, ignore minor variations, and avoid both overly specific and overly generic abstractions.

\par
Applying this rule generalization for all validated clusters produces our final AECR knowledge base, denoted as $\mathcal{K} = \{r_1, r_2, \dots, r_K\}$, where $r_k$ is an abstract event causal rule and $K$ denotes the total number of rules. 

\begin{figure}[t!]
	\centering
	\includegraphics[width=\columnwidth, trim=10 260 20 15, clip]{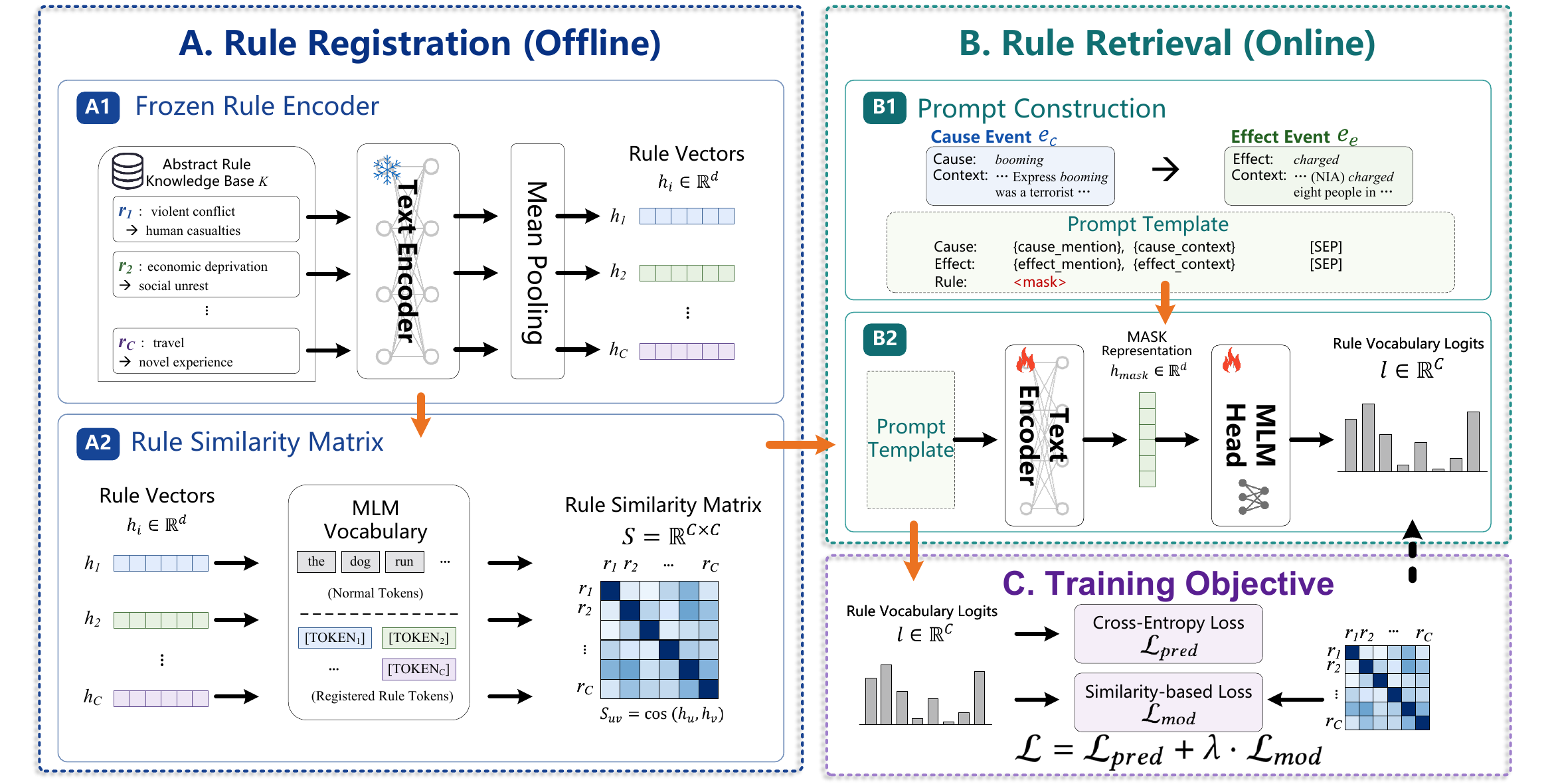}
	\caption{Overview of AECR knowledge base and retriever.}
	\label{Fig:Retriever}
    \vspace{-6pt}
\end{figure}
\subsection{Abstract Event Causal Rule Retriever}\label{SubSec:RuleRetriever}
To facilitate the utilization of the abstract causal rules for downstream tasks, we develop an AECR retriever that, given an instance of a causal event pair, retrieves the most explanatory abstract rules from the AECR knowledge base $\mathcal{K}$. To train this retriever, we construct training samples from the rule generation process described in Section~\ref{Subsec:AECRGeneration}: a causal event pair $p_i = (e_i^\texttt{c}, e_i^\texttt{e})$ is assigned an abstract rule $r_j$ if it falls into a valid rule cluster, and these qualified pairs $(p_i, r_i)$ serve as the data source for retriever training.

\par
$\bullet$ \textbf{Rule Registration.} A frozen text encoder (e.g., RoBERTa) is used for encoding each abstract rule \(r \in \mathcal{K}\) to output embeddings of all tokens in the rule text. We apply the mean pooling over these token embeddings to obtain a rule vector representation $\mathbf{h}_r$. For each rule, we create a dedicated virtual token, whose embedding is initialized with \(\mathbf{h}_r\) and fine-tuned during training. Additionally, we compute the cosine similarity between every pair of rule representations to obtain a similarity matrix $\mathbf{S} \in \mathbb{R}^{K \times K}$, where $\mathbf{S}_{u,v} = \cos(\mathbf{h}_u, \mathbf{h}_v)$.

\par
$\bullet$ \textbf{Rule Retrieval.} For a given instance of causal event pair $p_i = (e_i^\texttt{c}, e_i^\texttt{e})$, we construct the following prompt template:
\[
 \begin{aligned}
 &\texttt{Cause:}\ \text{mention}(e_i^\texttt{c}),\ \text{context}(e_i^\texttt{c}) \\
 &\texttt{[SEP] Effect:}\ \text{mention}(e_i^\texttt{e}),\ \text{context}(e_i^\texttt{e}) \\
 &\texttt{[SEP] Rule:}\ \langle\texttt{mask}\rangle
 \end{aligned}
\]
where $\text{mention}(e_i^\texttt{c})$ and $\text{mention}(e_i^\texttt{e})$ denote the event mentions, and $\text{context}(e_i^\texttt{c})$ and $\text{context}(e_i^\texttt{e})$ provide their sentential contexts. The template is fed into a trainable textual encoder. The hidden state $\mathbf{h}_{\texttt{mask}}$ of the $\langle\texttt{mask}\rangle$ token is then passed through the \textit{masked language model} (MLM) head. By slicing the vocabulary to retain only the virtual-token dimensions, we obtain the probability of each abstract rule being the correct match for the input causal pair:
\[
 P(r \mid p_i) = \text{Softmax}\left( \text{MLM}\left( \mathbf{h}_{\texttt{mask}} \right) \right)_{[\mathcal{V}]},
\]
where \([\mathcal{V}]\) denotes the slicing operation that retains only dimensions corresponding to virtual-tokens.

\par
$\bullet$ \textbf{Training Objective.} We adopt the standard cross-entropy loss with L2 regularization as the primary classification objective:
\[
    \mathcal{L}_{pred} = -\frac{1}{K} \sum_{k=1}^{K} \mathbf{y}^{(k)} \log\left(\hat{\mathbf{y}}^{(k)}\right) + \lambda \|\theta\|^2,
\]
where $K$ is the number of abstract event causal rules, $\mathbf{y}^{(k)}$ is the one-hot label, and $\hat{\mathbf{y}}^{(k)}$ is the predicted probability for the $k$-th rule.

\par
To exploit the semantic structure of the rule space, we introduce a complementary loss that relaxes penalties for semantically similar negative classes:
\[
    \mathcal{L}_{mod} = -\sum_{i=1}^{N} \log \frac{ \exp(z_{i, y_i} / \tau) }{ \exp(z_{i, y_i} / \tau) + \sum_{j \neq y_i} \exp\left( z_{i, j} / \tau - \mathbf{S}_{y_i, j} / \alpha \right) },
\]
where \(\tau\) is a temperature parameter, \(\alpha\) controls the influence of semantic similarities \(\mathbf{S}_{y_i, j}\), and \(z_{i,j}\) denotes the logit for class \(j\). This formulation preserves strict discrimination against semantically distant classes while avoiding over-penalization of confusions between near-synonymous rules.

The total training objective is:
\[
\mathcal{L} = \mathcal{L}_{pred} + \lambda \cdot \mathcal{L}_{mod}.
\]
After training, the AECR retriever parameters are frozen, enabling it to serve as a plug-and-play module for downstream applications.

%% file: Sections/ConstructionEvaluation.tex
\begin{table}[t]

    \centering

 \setlength{\tabcolsep}{3.5pt}
 \resizebox{\linewidth}{!}{
 \renewcommand{\arraystretch}{1}
 \begin{tabular}{@{}l|cccc|c}
 \toprule
 \multirow{2}{*}{\textbf{Datasets}} & \multicolumn{4}{c|}{\textbf{Average}} & \multirow{2}{*}{\textbf{ECGs}} \\ \cline{2-5}
 & \textbf{Nodes} & \textbf{Edges} & \textbf{Paths} & \textbf{Path Len.} & \\ \midrule
 MAVEN-CGEP & 8.4 & 12.9 & 2.7 & 5.0 & 5,308 \\
 ESC-CGEP & 11.0 & 24.9 & 4.9 & 4.9 & 363 \\ \bottomrule
 \end{tabular}
 }
\caption{Statistics of MAVEN-CGEP and ESC-CGEP dataset.}
\label{Tbl:DatasetStat}
\vspace{-25pt}
\end{table}


\section{Evaluation on AECR Construction}

\subsection{Datasets and AECR Knowledge Base}
We construct two AECR knowledge bases on two benchmark datasets: MAVEN-CGEP and ESC-CGEP~\cite{zhan2024would}. MAVEN-CGEP comprises 3,015 documents with 5,308 \textit{Event Causality Graphs} (ECGs) extracted from Wikipedia articles~\cite{wang2022maven}, covering a broad spectrum of open-domain event types such as political conflicts, natural disasters, and social movements. ESC-CGEP encompasses 243 documents (from news articles) with 363 ECGs~\cite{caselli2017event}, focusing on temporally grounded news storylines. Each dataset is organized as a collection of ECGs, where nodes represent events and directed edges denote causal relations. The detailed statistics of the two datasets are summarized in Table~\ref{Tbl:DatasetStat}.

\par
Applying our CACI framework described in Section~\ref{Sec:Construction} to the two datasets, we obtain two AECR knowledge bases, called MAVEN-AECR and ESC-AECR. Table~\ref{Tbl:AECRKBstat} summarizes the statistics of the intermediate products from the construction process. For MAVEN-CGEP, the CACI processes 32,263 original causal event pairs extracted from the ECGs. Among them, 6,179 pairs are filtered out through three mechanisms: (1) the Gatekeeper Agent discards causally implausible pairs, (2) the Abstraction Judge Agent rejects pairs whose abstractions fail to satisfy the quality criteria after the maximum retry attempts, and (3) during the clustering stage, concrete causal logics falling into clusters with fewer than $\mu$ members are removed, as such small clusters with very few samples represent weakly generalizable patterns. The remaining 26,084 pairs undergo abstraction and clustering, yielding 733 abstract event causal rules with an average cluster size of 35.6. For ESC-CGEP, from 6,875 original pairs, 844 are filtered out through these three mechanisms, and the remaining 6,031 pairs are consolidated into 180 abstract rules with an average cluster size of 33.5.

\begin{table}[htbp]
    \centering
	\setlength{\tabcolsep}{4pt} 
	\resizebox{\columnwidth}{!}{
		\renewcommand{\arraystretch}{1.1} 
	\begin{tabular}{@{}lcccccc}
		\toprule
		\multirow{2}{*}{\textbf{Dataset}} &
		\multirow{2}{*}{\makecell{\textbf{Original}\\\textbf{Pairs}}} &
		\multirow{2}{*}{\makecell{\textbf{Filtered}\\\textbf{Pairs}}} &
		\multirow{2}{*}{\makecell{\textbf{Retained}\\\textbf{Pairs}}} &
		\multirow{2}{*}{\makecell{\textbf{Abstract}\\\textbf{Rules}}} &
		\multicolumn{2}{c}{\textbf{Cluster Size}} \\
		\cmidrule(l){6-7}
		& & & & & \textbf{Avg.} & \textbf{Median} \\
		\midrule
		
		\textbf{MAVEN} & 32,263 & 6,179 & 26,084 & 733 & 35.6 & 15 \\
		\textbf{ESC}   & 6,875 & 844  & 6,031 & 180  & 33.5 & 15.5 \\
		
		\bottomrule
	\end{tabular}
	}
	\caption{AECR Construction Statistics by our CACI framework on the MAVEN-CGEP and ESC-CGEP datasets.}
	\label{Tbl:AECRKBstat}
    \vspace{-10pt}
\end{table}


\subsection{Human Evaluation of AECR Quality}
We performed human evaluation of the extracted abstract causal rules with 14 graduate student annotators. The AECR construction is evaluated from three perspectives. (1) Reasonableness: The degree to which extracted causal rules are logically valid in human evaluation.
(2) Discriminability: The degree to which concrete causal event pairs map to abstract causal rules.
(3) Usability: The degree to which concrete causal event pairs deserve abstraction into causal rules.


\begin{table}[htbp]
  \centering
  \renewcommand{\arraystretch}{1.1}
  \resizebox{\linewidth}{!}{
  \begin{tabular}{lccccc}
    \hline
    \textbf{Dataset} & \textbf{Avg.} & \textbf{Std.} & \textbf{Median} & \textbf{PHC(\%)} & \textbf{Agreement($\alpha$)} \\
    \hline
    \textbf{MAVEN-AECR} & 4.57 & 0.78 & 5.00 & 91\% & 0.82 \\
    \textbf{ESC-AECR} & 4.56 & 0.73 & 5.00 & 88\% & 0.87 \\
    \hline
  \end{tabular}}
  \caption{Human evaluation results on rule reasonableness.}
  \label{tab:reliability}
  \vspace{-10pt}
\end{table}


\par
$\bullet$ \textbf{Reasonableness Evaluation.}  To conduct the reasonableness assessment, we randomly sampled 200 abstract causal rules from each AECR knowledge base. Three independent annotators were tasked with rating the reasonableness of these rules on a 5-point Likert scale: Given an event pair and its corresponding extracted abstract rule, the annotators scored how reasonably the rule could be generalized from the provided event pair.

\par
Table~\ref{tab:reliability} reports the credibility evaluation results. These randomly sampled rules from both knowledge bases yield high average scores, with a median score of 5.00. This demonstrates that most rules deliver distinct deterministic causality. The Proportion of High-Reliability Candidates (PHC) refers to the percentage of rules scoring 4–5. A high PHC value implies a very low ratio of spurious causal rules, demonstrating that the extracted causal rules can serve as trustworthy prior knowledge. Note that the validity of our manual evaluation is supported by high inter-annotator agreement, with Krippendorff’s $\alpha$ attaining 0.82 and 0.87 respectively.

\begin{table}[htbp]
	\centering
	
	\setlength{\tabcolsep}{4pt} 
	\resizebox{\columnwidth}{!}{
		\renewcommand{\arraystretch}{1}
		\begin{tabular}{llcccc}
			\toprule
			\multirow{2}{*}{\textbf{Dataset}} & \multirow{2}{*}{\textbf{Setting}} & \multirow{2}{*}{\textbf{\# Samp.}} & \multicolumn{2}{c}{\textbf{Accuracy (\%)}} & \multirow{2}{*}{\textbf{Agreement ($\alpha$)}} \\
			\cmidrule(lr){4-5}
			& & & \textbf{Maj.} & \textbf{All} & \\
			\midrule
			
			\multirow{2}{*}{\textbf{MAVEN-AECR}}
			& Normal Neg. & 100 & 99.0 & 93.0 & 0.93 \\
			& Hard Neg.   & 100 & 98.0 & 76.0 & 0.77 \\
			\midrule
			
			\multirow{2}{*}{\textbf{ESC-AECR}}
			& Normal Neg. & 100 & 100.0 & 90.0 & 0.91 \\
			& Hard Neg.   & 100 & 99.0 & 50.0 & 0.55 \\
			\bottomrule
		\end{tabular}
	}
	\caption{Human evaluation results on rule discriminability. Accuracy is reported for Majority Vote (Maj.) and All Correct (All). $\alpha$ denotes the inter-annotator agreement.}
    \label{tab:discriminability}
    \vspace{-10pt}
\end{table}


\par
$\bullet$ \textbf{Discriminability Evaluation.} We design a single-choice human evaluation task to quantify the discriminability of extracted abstract causal rules against concrete event pairs. Given an event pair, we assemble its corresponding extracted rule alongside three alternative rules to form a single-choice question. Two assessment configurations are adopted: Normal Negative uses three random irrelevant rules as distractors, and Hard Negative includes an extra highly analogous causal rule as a distractor.

\par
Table~\ref{tab:discriminability} reports the discriminability evaluation results. Under the Normal Negative configuration, majority vote accuracy reaches 99.0\% on MAVEN-AECR and 100.0\% on ESC-AECR, with Krippendorff’s $\alpha$ values of 0.93 and 0.91 respectively. These results indicate near-perfect inter-annotator agreement, demonstrating that abstract rules faithfully capture concrete causal semantics and exhibit strong discriminability. Under the Hard Negative configuration, majority vote accuracy stands at 98.0\% on MAVEN-AECR and 99.0\% on ESC-AECR. This demonstrates that the extracted rules retain strong discriminability against highly analogous causal rules. Both full-match accuracy and inter-annotator Krippendorff’s $\alpha$ decrease substantially for the ESC-AECR, falling to 50.0\% and 0.55 respectively. The root cause is the dense, fine-grained causal semantics within ESC-AECR: annotators struggle to distinguish subtle differences among highly similar rules, which leads to inconsistent annotations. Nevertheless, the consistently high majority-vote accuracy reveals that annotators as a group still prefer the ground-truth rule as the most appropriate match.
  
\begin{figure}[htbp]  
		\centering
		\includegraphics[width=\columnwidth]{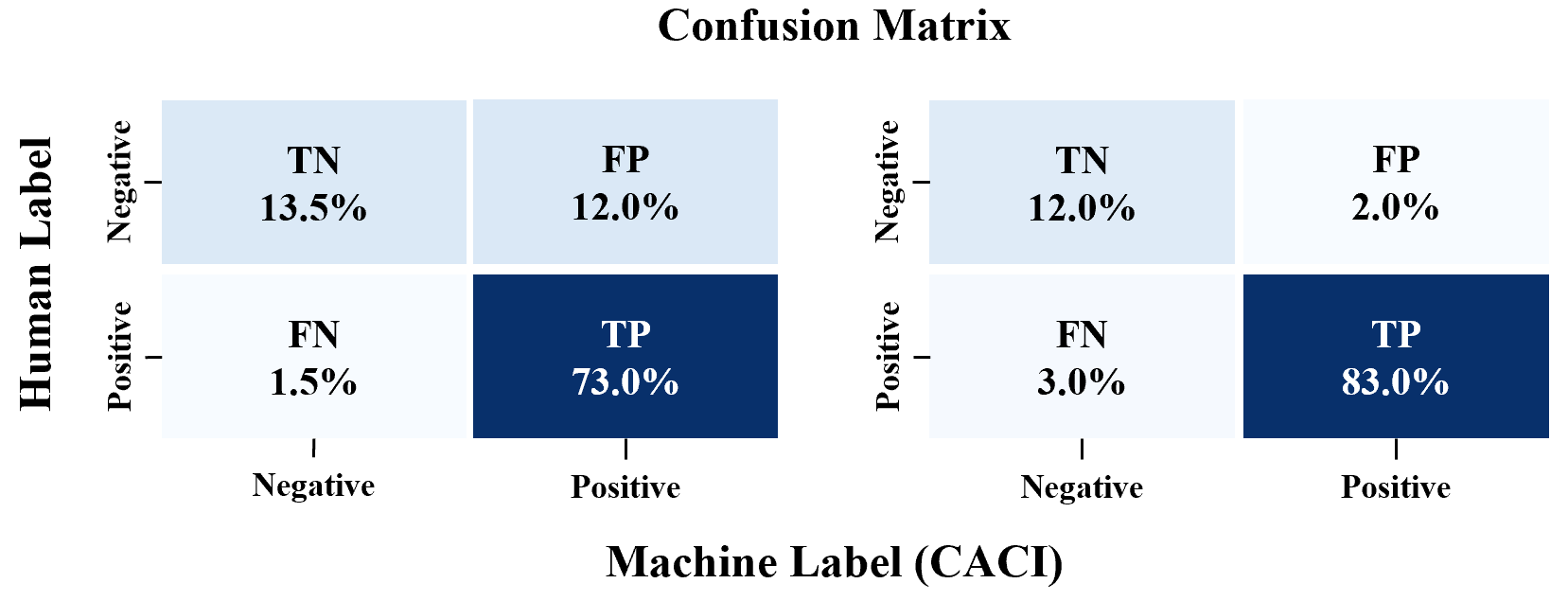}
		\caption{Confusion matrix of human and machine evaluation on rule usability. (Left) MAVEN; (Right) ESC.}
		\label{Fig:usability}  
\end{figure}

\par    
$\bullet$ \textbf{Usability Evaluation.} Recall that our AECR construction pipeline filters out certain concrete event pairs to exclude them from rule generation. We evaluate whether this filtering mechanism effectively distinguishes event pairs with high utility for rule generalization. For each dataset, we sample 170 concrete event pairs retained for the subsequent rule generation phase and another 30 filtered-out pairs as controls. Each annotator independently assigns a binary \textsf{YES}/\textsf{NO} label to each sample to judge its usability in deriving abstract rules, with the final consensus label determined via majority voting to resolve individual divergence.

\par
Figure~\ref{Fig:usability} presents the confusion matrices from human evaluation of event pair usability. Our construction pipeline yields low false negative (FN) rates: FN = 1.5\% for MAVEN and 3.0\% for ESC, with recall scores as high as 98.0\% (MAVEN) and 96.5\% (ESC). This indicates that the filtering mechanism rarely discards usable causal pairs, avoiding the loss of rare long-tail causal rules in the early filtering stage. Furthermore, our pipeline attains precision of 85.9\% on MAVEN and 97.6\% on ESC by filtering out unusable and noisy event pairs. Overall, our pipeline achieves a favorable balance: it preserves causal rule structures at high recall and leverages an effective filtering mechanism to mitigate model hallucinations.

\subsection{Evaluation of AECR Retriever}
The constructed AECR knowledge base is built to act as an external knowledge source for downstream applications, yet its practical utility hinges on a well-performing AECR retriever. That is, given a concrete causal event pair, the AECR retriever determines its most relevant abstract causal rules from the knowledge base. To train and evaluate the retriever, we take the concrete event pairs together with their corresponding rules in each knowledge base and split them into training, validation, and test sets at an 8:1:1 ratio. The retriever is trained independently on each knowledge base (c.f., Section~\ref{SubSec:RuleRetriever}), and we report the test-set performance of the checkpoint that performs best on the validation set.

\par
We compare our AECR retriever with two commonly used retrievers: (1) Sentence-BERT retriever~\cite{reimers2019sentence} (SBERT retriever), which independently encodes a concrete event pair and all abstract rules into embeddings and retrieves rules based on the cosine similarity. (2) LLM-based Re-ranker (LLM Reranker), which first employs the SBERT retriever to recall a candidate set of the top-100 most similar rules, and then utilizes an LLM (the llama-3.1-8b~\cite{grattafiori2024llama3herdmodels} in our experiments) to perform a single-round, instruction-based re-ranking to select the top-10 most causally relevant rules from this candidate pool.

\begin{table}[thbp]
\centering
\setlength{\tabcolsep}{5pt}
\resizebox{\columnwidth}{!}{
\renewcommand{\arraystretch}{1}
\begin{tabular}{@{}llcccc}
\toprule
\textbf{Dataset} & \textbf{Retriever} & \textbf{MRR} & \textbf{Hit@1} & \textbf{Hit@3} & \textbf{Hit@10} \\
\midrule
\multirow{3}{*}{MAVEN-AECR}
 & SBERT retriever& 14.39 & 5.14 & 13.72 & 37.06 \\
 & LLM Re-ranker & 16.17 & 7.40 & 20.31 & 39.82 \\
 & Our retriever & \cellcolor[HTML]{e0f0ff}\textbf{52.64} & \cellcolor[HTML]{e0f0ff}\textbf{40.55} & \cellcolor[HTML]{e0f0ff}\textbf{59.87} & \cellcolor[HTML]{e0f0ff}\textbf{75.05}\\
\midrule
\multirow{3}{*}{ESC-AECR}
 & SBERT retriever & 27.40 & 12.46 & 32.06 & 60.63 \\
 & LLM Re-ranker & 32.19 & 13.29 & 36.38 & 56.48 \\
 & Our retriever & \cellcolor[HTML]{e0f0ff}\textbf{65.98} & \cellcolor[HTML]{e0f0ff}\textbf{51.50} & \cellcolor[HTML]{e0f0ff}\textbf{76.91} & \cellcolor[HTML]{e0f0ff}\textbf{90.03}\\
\bottomrule
\end{tabular}
}
\caption{Evaluation on AECR retriever performance.}\label{Tbl:RetrieverPerformance}
\label{tab:retrieval}
\vspace{-10pt}
\end{table}
\par
Table~\ref{Tbl:RetrieverPerformance} compares the top-10 retrieval performance across the three retrievers. It is not unexpected that our retriever outperforms the other two, since it is specially trained on our constructed AECR knowledge base. The SBERT retriever, which relies solely on semantic similarity, fails to capture deep logical equivalence between concrete event pairs and their corresponding abstract causal rules. The LLM-based reranker achieves moderate improvements over SBERT by drawing on its internal parametric knowledge. Nevertheless, the performance of our AECR retriever remains suboptimal, as reflected by its low Hit@1 score. We leave the design of more powerful AECR retrievers as future work.

%% file: Sections/Application.tex
\begin{figure*}[t!]  
	\centering
	\includegraphics[width=\textwidth, trim=10 250 10 15, clip]{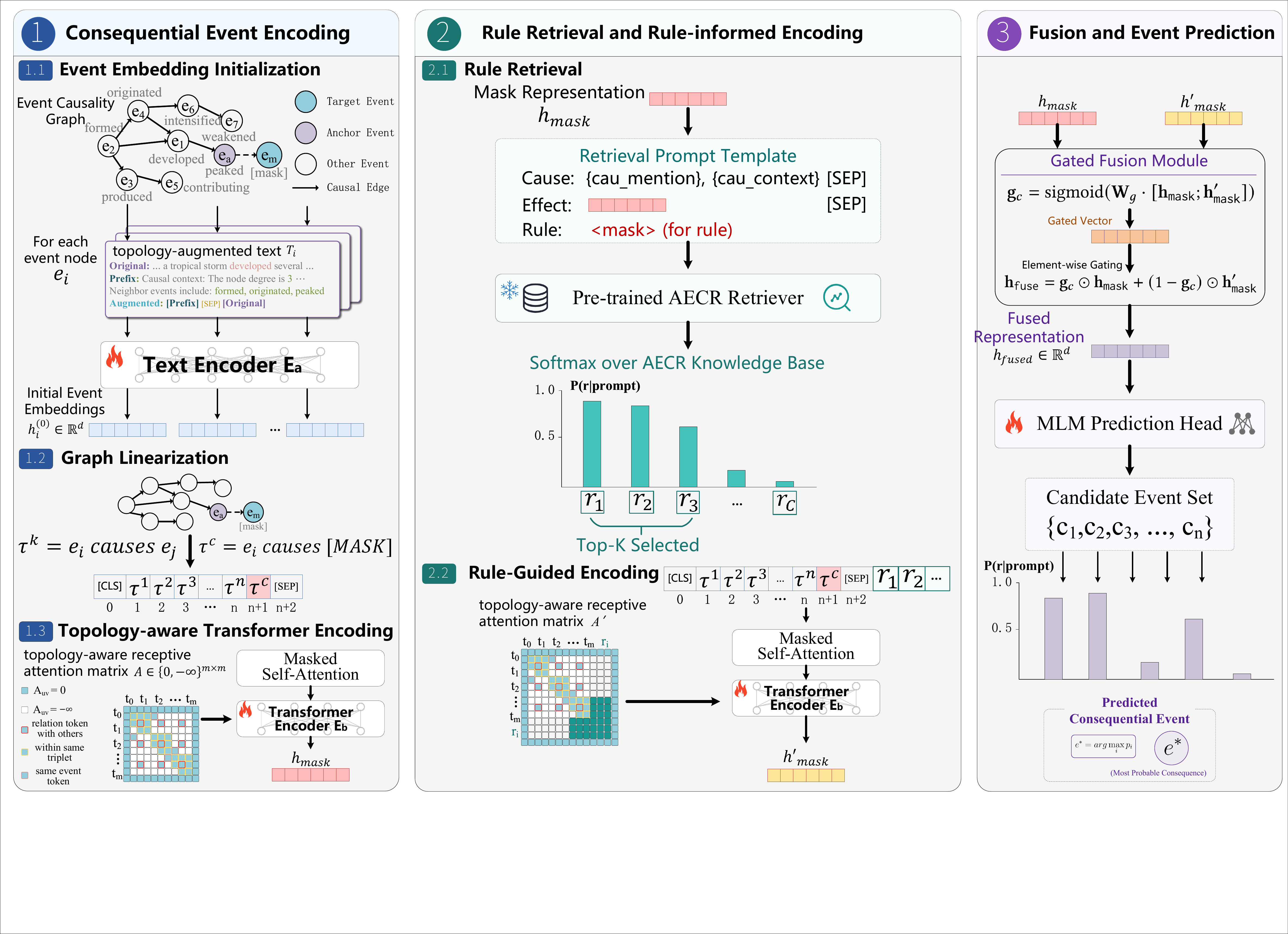}
	\caption{Overview of AR-GCAE Framework for Event Prediction.}
	\label{Fig:CGEP}
\end{figure*}

\section{Application of AECR on Event Prediction}
This section details the application of our constructed AECR knowledge base $\mathcal{K}$ to the \textit{causality graph event prediction} (CGEP) task. 

\par
$\bullet$ \textbf{Problem Formulation.} The CGEP task aims to predict the consequential event for an \textit{anchor event} on an \textit{event causality graph} (ECG). Specifically, an ECG is a directed acyclic graph $\mathcal{G} = (\mathcal{E}, \mathcal{R})$, where the node set $\mathcal{E}$ represents historical events, and the edge set $\mathcal{R}$ denotes the directed causal relations between event nodes. Each event node $e_i \in \mathcal{E}$ is a tuple $e_i = (m_i, s_i)$, where $m_i$ is an event mention (a word or phrase) and $s_i$ is the sentence containing $m_i$, providing its contextual background.  Given an anchor event node $e_a \in \mathcal{E}$ and a candidate event set $\mathcal{E}_c$, the objective is to identify the most probable consequential event $e^* \in \mathcal{E}_c$ (to be) caused by $e_a$.

\par
We propose an \textit{Abstract Rule-Guided Causal Attention Encoder} (AR-GCAE)  that first retrieves relevant abstract causal rules from $\mathcal{K}$ and then injects them into the attention-based graph encoding process, providing causal guidance for the final event prediction. 

\subsection{Consequential Event Encoding}\label{SubSec: ECPE}
This module encodes the entire event causality graph to obtain a representation $\mathbf{h}_\texttt{mask}$ of the unknown consequential event for the query pair $\langle e_a, \texttt{[MASK]}\rangle$. The process consists of three steps: event embedding initialization, graph linearization, and topology‑aware causal pair encoding.

\par
$\bullet$ \textbf{Event Embedding Initialization.} For each event node $e_i \in \mathcal{E}$, we construct a semi-structured prefix that encapsulates the 1-hop local topology centered at $e_i$, including its node degree and the mentions of its neighbors, while excluding the target consequential event to prevent data leakage. This prefix is concatenated with the event sentence $s_i$ containing the event mention $m_i$ (delimited by a $\langle \texttt{[SEP]}\rangle$ token) to form a topology-augmented text $\textsf{T}_i$, which is then fed into a text encoder $\mathbb{E}_a$ (e.g., the RoBERTa in our implementation) to obtain the initial event embedding $\mathbf{h}_i^{(0)}$.

\par
$\bullet$ \textbf{Graph Linearization.} We represent each causal event pair $\langle e_i, e_j\rangle \in\mathcal{G}$ on the ECG as a triplet $\tau$:
\[ 
    \tau\ =\ m_i\ causes\ m_j,
\]
where $m_i$ and $m_j$ are the event mention of the cause event and effect event, respectively. The prediction target is formulated as a query triplet 
\[
    \tau^c\ = m_a\ causes\ \texttt{[MASK]}
\]
where $m_a$ is the event mention of the anchor event $e_a$. Following \cite{zhan2024would}, all triplets (including $\tau^c$) are sorted in descending order of their shortest path distance from the involved event nodes to the anchor node. The sorted triplets are then concatenated to form a graph prompt 
$\mathcal{T}$ 
\[ 
    \mathcal{T}\ =\ \texttt{[CLS] } \tau_1\ \tau_2\ \cdots\ \tau_n\ \tau^c\ \texttt{[SEP]},
\]
where \texttt{[CLS]} and \texttt{[SEP]} are special tokens marking the beginning and end of the sequence.

\par
$\bullet$ \textbf{Reception Matrix Definition.} We define a \textit{topology‑aware receptive attention matrix} $\mathbf{A}\in\{0, -\infty\}^{m\times m}$, where $m$ is the number of tokens in $\mathcal{T}$. For a token $t_j$ in $\mathcal{T}$, its allowed interactions with other tokens $t_i$ are determined as follows: If $t_j$ is a special token (e.g., \texttt{[CLS]}, \texttt{[SEP]}, or \texttt{[MASK]}), it attends to all tokens. If $t_j$ is a relation token (i.e. “causes”), it attends only to the cause and effect event tokens within the same triplet (i.e., $t_{j-1}$ and $t_{j+1}$) and to all relation tokens across $\mathcal{T}$. If $t_j$ is an event token, it attends to the cause event and relation token within the same triplet, and to other occurrences of the same event token elsewhere in the sequence; Moreover, a cause event may attend to its effect event, but not vice versa. Correspondingly, we set $\mathbf{A}_{uv}=0$ if the token $u$ is allowed to attend to the token $v$, and $\mathbf{A}_{uv}=-\infty$ otherwise. This masking mechanism ensures that each token only aggregates information from topologically relevant tokens.

\par
$\bullet$ \textbf{Topology-aware Transformer Encoding.} We initialize the embeddings of event tokens in $\mathcal{T}$ using the pre‑computed $\mathbf{h}_i^{(0)}$. Embeddings of relation tokens and special tokens are directly looked up from standard embedding matrix. The entire sequence is then fed into the Transformer $\mathbb{E}_b$. The self‑attention computation at layer $l$ is modified as:
\[
    \mathbf{H}^{l+1}=\text{softmax}\left(\frac{1}{{\sqrt{d}}} (\mathbf{H}^l\mathbf{W}^l_q)(\mathbf{H}^l\mathbf{W}^l_k)^\textsf{T}+\mathbf{A}\right)
    (\mathbf{H}^l\mathbf{W}^l_v),
\]
where $\textbf{W}^l_q, \textbf{W}^l_k, \textbf{W}^l_v$ are learnable projections and $d$ is the hidden dimension.

\par
After the last Transformer layer, we obtain the hidden state of the \texttt{[MASK]} token, denoted as $\mathbf{h}_\texttt{mask}$, which serves as the representation of the unknown consequential event for the query pair $\langle e_a, \texttt{[MASK]}\rangle$.

\subsection{Rule Retrieval and Rule-Informed Encoding}
To obtain a rule-informed representation of the \texttt{[MASK]} token for event prediction, we retrieve relevant abstract event causal rules for the query pair $\langle e_a, \texttt{[MASK]} \rangle$ and incorporate them into the graph encoding process. 

\par
$\bullet$ \textbf{Rule Retrieval.} For the query causal pair \(\langle e_a, \texttt{[MASK]} \rangle\), we obtain the $\texttt{[MASK]}$ token representation $\mathbf{h}_{\texttt{mask}}$ from the topology-aware encoder $\mathbb{E}_b$ (Section~\ref{SubSec: ECPE}). This representation subsumes the contextual semantics of the anchor event $e_a$ and the graph topology, serving as a latent proxy for the unknown effect event. We construct the retrieval prompt as follows:
\[
    \begin{aligned}
        &\texttt{Cause:}\ \text{mention}(e_a),\ \text{context}(e_a) \\
        &\texttt{[SEP] Effect:}\ \texttt{[PH]} \\
        &\texttt{[SEP] Rule:}\ \texttt{[MASK]}
    \end{aligned}
\]
where $\texttt{[PH]}$ is a placeholder token. After tokenization, all tokens are initialized with their standard word embeddings, except for $\texttt{[PH]}$, whose embedding is directly replaced with $\mathbf{h}_{\texttt{mask}}$. The resulting mixed embedding sequence is then fed into the pre-trained retriever (Section~\ref{SubSec:RuleRetriever}), which outputs a probability distribution over the rule set. We select the top-K rules with the highest probabilities as $\mathcal{R}_{retrieved}=\{r_1, \dots, r_K\}$

\par
$\bullet$ \textbf{Rule-Guided Encoding.} Given the selected $K$ rules, we form an augmented graph prompt $\mathcal{T}^\prime$ by appending these rule texts to the original graph prompt $\mathcal{T}$:
\[
    \mathcal{T}^{\prime} = \mathcal{T} \oplus [\texttt{SEP}] \oplus \texttt{[Rule]}: \text{desc}(r_1),\ \ldots,\ \text{desc}(r_K),
\]
where $\oplus$ denotes concatenation and $\texttt{[Rule]}$ marks the beginning of the rule section. The attention matrix $\mathbf{A}^\prime$ for $\mathcal{T}^\prime$ follows the same topology-aware rules as $\mathbf{A}$ (Section~\ref{SubSec: ECPE}) for tokens within the original graph prompt. For the rule section, we allow bidirectional attention between the query triplet $\tau^c$ and all rule tokens; ordinary triplets that do not contain the $\texttt{[MASK]}$ token are not allowed to attend to rule tokens, nor vice versa. Special tokens retain full visibility over the entire sequence.

\par
We then encode $\mathcal{T}^\prime$ with the same text encoder $\mathbb{E}_b$ used in Section~\ref{SubSec: ECPE}, reusing the pre-computed initial event embeddings $\mathbf{h}_i^{(0)}$ for event tokens. Embeddings for relation tokens, special tokens, and rule texts are looked up from the standard embedding matrix. The self-attention computation uses the attention matrix $\mathbf{A}^\prime$. After the last layer of the encoder $\mathbb{E}_b$, we obtain the updated hidden state of the $\texttt{[MASK]}$ token, denoted as $\mathbf{h}^\prime_{\texttt{mask}}$.

\subsection{Fusion and Event Prediction}
We adopt a gated attention mechanism to combine the $\texttt{[MASK]}$ representations from the two passes, $\mathbf{h}_{\texttt{mask}}$ from the first pass without using rules and $\mathbf{h}^\prime_{\texttt{mask}}$ from the second pass with rule guidance:
\[ 
    \mathbf{g}_c = \text{sigmoid}(\mathbf{W}_g\cdot[\mathbf{h}_{\texttt{mask}};\mathbf{h}^\prime_{\texttt{mask}}]),
\]
\[
    \mathbf{h}_{\texttt{fuse}}=\mathbf{g}_c\odot\mathbf{h}_{\texttt{mask}} + (1-\mathbf{g}_c)\odot\mathbf{h}_{\texttt{mask}}^\prime,
\]
\[
    \text{Prob}\left(\texttt{[MASK]}\right)=\text{MLMHead}\left(\mathbf{h}_{\texttt{fuse}}\right),
\]
where $\mathbf{W}_g$ is a trainable parameter matrix. The fused representation $\mathbf{h}_\texttt{fuse}$ is then fed into an MLM head to produce a probability distribution over the candidate event set $\mathcal{E}_c$.

\par
\textbf{Training Objective.}
The primary prediction loss $\mathcal{L}_{\text{pred}}$ is the cross-entropy between the predicted distribution and the ground-truth event:
\[
    \mathcal{L}_{\text{pred}} = -\frac{1}{N}\sum_{i=1}^{N}\log p(y_i \mid \mathbf{h}_{\texttt{fuse}}^{(i)}),
\]
where $N$ is the number of training samples, $y_i$ is the ground-truth event for the $i$-th instance, and $\mathbf{h}_{\texttt{fuse}}^{(i)}$ is the fused representation from the $i$-th sample. The entire model is trained end-to-end using AdamW optimizer.

%% file: Sections/ApplicationExperiment.tex
\section{Experiments on AECR Application}


\subsection{Experimental Setup}

\textbf{Datasets.}
We conduct downstream event prediction experiments on the two benchmarks introduced in Section~\ref{Sec:Construction}, MAVEN-CGEP and ESC-CGEP. We adopt the same partitioning protocol as prior work~\cite{zhan2024would} to ensure a fair comparison. For MAVEN-CGEP, the original development set is repurposed as the test set, and a new development set is drawn by randomly holding out 20\% of the original training instances. For ESC-CGEP, whose scale is considerably smaller, we set aside the last two topics as the development set and run 5-fold cross-validation over the remaining twenty topics, reporting the averaged results.

\par
\textbf{Candidate Set Construction.}
For each instance, the model is asked to rank a candidate set of events. Following SeDGPL~\cite{zhan2024would}, this set consists of the ground-truth consequential event together with a number of negative candidates sampled uniformly from the tail nodes (i.e., events with no outgoing causal edge) of other ECGs. Since the negatives are drawn per instance, the candidate set differs across instances, while its size is fixed to 512 candidates on MAVEN-CGEP and 256 on ESC-CGEP. The candidate sets are shared identically across all compared methods.

\par
\textbf{Competitors.}
We compare \textsf{AR-GCAE} against two groups of competitors: encoder-based and LLM-based approaches. The \textit{encoder-based} group learns event representations over the causality graph with pretrained encoders. \textsf{SEP-BART}~\cite{zhu2023generative} adopts a generative objective to directly produce the subsequent event, while the \textsf{MCPredictor}~\cite{bai2021integrating} encodes narrative event chains with a Transformer. The \textsf{SeDGPL}~\cite{zhan2024would} designs graph prompts that integrate event types with contextual information, and \textsf{CBLiP}~\cite{dutta2025replacing} introduces a connection-biased attention matrix to capture structural causal dependencies. \textsf{SEDA}~\cite{zhan2025} constructs a multi-faceted event graph to model complex evolution patterns, and \textsf{TRACE}~\cite{zheng2026} further augments the causality graph with LLM-generated nodes and edges and applies a robust, position-aware Transformer encoder to counter structural deficiency. The \textit{LLM-based} group probes general-purpose \textsf{Llama-3.1-8B}~\cite{grattafiori2024llama3herdmodels} and \textsf{GPT-3.5-turbo}~\cite{ouyang2022training}. Following prior work~\cite{zhan2024would, zhan2025, zheng2026}, we evaluate them under zero-shot and $3$-shot ($\Delta$) settings, and additionally equip each with a decompositional chain-of-thought (CoT) pipeline that progressively reasons over event structure and context before ranking the shared candidate set.

\par
\textbf{Implementation Details.}
Our \textsf{AR-GCAE} is built on the 768-dimensional RoBERTa encoder from the HuggingFace Transformers library, and trained with PyTorch and CUDA on NVIDIA RTX 3090 GPUs. We use a batch size of $1$ and inject the top $K=3$ retrieved abstract rules for each dataset. On MAVEN-CGEP, we train for $10$ epochs with a learning rate of $1\mathrm{e}{-6}$; on ESC-CGEP, we train for $15$ epochs with a learning rate of $5\mathrm{e}{-6}$. All hyperparameters are tuned according to the performance on the development set.

\par
\textbf{Evaluation Metrics.}
Consistent with prior CGEP studies~\cite{zhan2024would, zhan2025, zheng2026}, we report Mean Reciprocal Rank (MRR) and Hit@$n$ with $n\in\{1,3,10,20,50\}$. MRR averages the reciprocal rank assigned to the ground-truth event, while Hit@$n$ measures the fraction of instances whose ground-truth event is ranked within the top $n$ candidates; for both, higher values indicate better performance. To assess whether the improvements over the state-of-the-art methods are statistically meaningful, we also conduct a paired one-tailed permutation test with Bonferroni correction.

\begin{table*}[htbp]
		\renewcommand{\arraystretch}{1} 
		\resizebox{\linewidth}{!}{
			\begin{tabular}{@{}l|cccccc|cccccc} 
				\toprule
				\midrule
				& \multicolumn{6}{c|}{\cellcolor[HTML]{f7f9f1}\textbf{MAVEN-CGEP}} & \multicolumn{6}{c}{\cellcolor[HTML]{f7f9f1}\textbf{ESC-CGEP}} \\
				\cmidrule(lr){2-7} \cmidrule(lr){8-13} 
				\multirow{-2}{*}{\textbf{Model}} &
				\cellcolor[HTML]{fef8e6}MRR & \cellcolor[HTML]{fef8e6}Hit@1 & \cellcolor[HTML]{fef8e6}Hit@3 & \cellcolor[HTML]{fef8e6}Hit@10 & \cellcolor[HTML]{fef8e6}Hit@20 & \cellcolor[HTML]{fef8e6}Hit@50 & \cellcolor[HTML]{fef8e6}MRR & \cellcolor[HTML]{fef8e6}Hit@1 & \cellcolor[HTML]{fef8e6}Hit@3 & \cellcolor[HTML]{fef8e6}Hit@10 & \cellcolor[HTML]{fef8e6}Hit@20 & \cellcolor[HTML]{fef8e6}Hit@50 \\ \midrule
				\textsf{SEP-BART} &
				24.7 & 19.5 & 24.5 & 34.8 & 42.6 & 53.6 & 16.0 & 12.5 & 16.8 & 21.1 & 28.6 & 38.9 \\
				\textsf{MCPredictor} &
				18.1 & 13.0 & 18.4 & 27.3 & 32.0 & 43.2 & 9.7 & 8.4 & 10.9 & 17.4 & 22.2 & 37.5\\
				\textsf{SeDGPL} &
				27.9 & 21.9 & 28.9 & 40.8 & 48.1 & 57.9 & 19.6 & 15.2 & 18.1 & 22.3 & 29.9 & 41.9 \\
				\textsf{CBLiP} & 28.2 & 22.3 & 28.3 & 38.6 & 46.4 & 55.4 & 16.8 & 13.6 & 16.4 & 19.9 & 28.9 & 41.1 \\
				\textsf{SEDA} & 30.4 & 25.3 & 30.5 & 40.9 & 48.5 & 58.3 & 20.0 & \cellcolor[HTML]{f0f7ff}\underline{16.3} & 18.4 & 27.3 & 34.4 & 43.3 \\
				\textsf{TRACE} & \cellcolor[HTML]{f0f7ff}\underline{36.2} & \cellcolor[HTML]{f0f7ff}\underline{28.9} & \cellcolor[HTML]{f0f7ff}\underline{37.7} & \cellcolor[HTML]{f0f7ff}\underline{50.6} & \cellcolor[HTML]{f0f7ff}\underline{59.4} & \cellcolor[HTML]{f0f7ff}\underline{71.3} & \cellcolor[HTML]{f0f7ff}\underline{20.8} & 15.2 & \cellcolor[HTML]{f0f7ff}\underline{20.2} & \cellcolor[HTML]{f0f7ff}\underline{30.7} & \cellcolor[HTML]{e0f0ff}\textbf{43.4} & \cellcolor[HTML]{e0f0ff}\textbf{59.8} \\
				\midrule
				\textsf{Llama-3.1-8B} &
				9.6 & 5.0 & 11.1 & 20.2 & 24.5 & 26.6 & 6.7 & 1.1 & 8.9 & 20.2 & 26.3 & 29.2 \\
				\textsf{GPT-3.5-turbo} &
				14.6 & 8.1 & 17.1 & 28.1 & 33.3 & 39.5 & 10.1 & 4.9 & 11.4 & 20.5 & 25.2 & 31.5 \\
				$\textsf{Llama-3.1-8B}^\Delta$ &
				10.8 & 5.4 & 14.9 & 21.8 & 25.1 & 34.6 & 8.3 & 0.3 & 12.6 & 21.2 & 28.8 & 34.9 \\
				$\textsf{GPT-3.5-turbo}^\Delta$ &
				14.8 & 10.4 & 16.4 & 28.6 & 34.0 & 41.7 & 10.6 & 5.9 & 12.9 & 20.7 & 25.9 & 34.3 \\
				$\textsf{Llama-3.1-8B + CoT}$ &
				19.1 & 8.1 & 23.3 & 44.5 & 50.2 & 55.1 & 17.5 & 4.9 & 18.6 & 24.8 & 41.4 & 47.3 \\
				\textsf{GPT-3.5-turbo + CoT} &
				17.8 & 9.5 & 19.1 & 34.9 & 47.8 & 57.2 & 12.7 & 7.3 & 15.8 & 22.2 & 25.1 & 28.6 \\ \midrule
				${\textsf{Our AR-GCAE}}$ &
				\cellcolor[HTML]{e0f0ff}\textbf{40.5} & \cellcolor[HTML]{e0f0ff}\textbf{32.1} & \cellcolor[HTML]{e0f0ff}\textbf{42.4} & \cellcolor[HTML]{e0f0ff}\textbf{57.4} & \cellcolor[HTML]{e0f0ff}\textbf{65.8} & \cellcolor[HTML]{e0f0ff}\textbf{76.0} & \cellcolor[HTML]{e0f0ff}\textbf{22.8} & \cellcolor[HTML]{e0f0ff}\textbf{17.3} & \cellcolor[HTML]{e0f0ff}\textbf{22.0} & \cellcolor[HTML]{e0f0ff}\textbf{32.4} & \cellcolor[HTML]{f0f7ff}\underline{43.1} & \cellcolor[HTML]{f0f7ff}\underline{58.6} \\ \midrule
				\bottomrule
			\end{tabular}
		}

		\caption{Comparison of overall event prediction performance (\%) on the MAVEN-CGEP and ESC-CGEP datasets.}
		\label{tab:main_result}
	\end{table*}
\subsection{Main Results}
\textbf{Overall Results.}
Table~\ref{tab:main_result} reports the overall comparison between \textsf{AR-GCAE} and all competitors on the MAVEN- and ESC-CGEP benchmarks. Our \textsf{AR-GCAE} consistently surpasses the competitors on both datasets, with an especially clear advantage on the top-ranked metrics that matter most for prediction. More concretely, on the larger and more challenging MAVEN-CGEP, \textsf{AR-GCAE} lifts MRR(\%) from 36.2 to 40.5 and Hit@1(\%) from 28.9 to 32.1 over the strongest competitor \textsf{TRACE}, corresponding to relative gains of $11.9\%$ and $11.1\%$, both statistically significant ($p<0.05$) under the paired one-tailed permutation test with Bonferroni correction. These results validate the effectiveness of grounding consequential event prediction in abstract event causal rules.

\par
\textbf{Comparison with Causal-Encoder Baselines.}
The first group of competitors devises dedicated causal encoders that learn event representations directly over the event causality graph. \textsf{SeDGPL}, \textsf{SEDA}, and \textsf{TRACE} progressively strengthen this line by coupling graph prompt learning with distance- or position-aware linearization and by mitigating structural noise during training. Nevertheless, all of them reason exclusively over the concrete events and edges observed in the training graphs, so the causal knowledge they acquire remains bound to instance-level co-occurrence patterns of specific event mentions. \textsf{AR-GCAE} departs from this paradigm by retrieving relevant AECRs, each describing how the parent-concept event of the cause instance can trigger another parent-concept event of an effect instance (i.e., relation-level other than instance-level causal knowledge), and by fusing this relation-level causal knowledge into the graph encoding process. This supplies a layer of generalizable causal knowledge that pure graph topology modeling cannot recover from the observed graph alone, which we regard as the principal reason for the performance gains over even the strongest baseline in this group.

\par
\textbf{Comparison with LLM Baselines.}
The second group probes general-purpose LLMs under zero-shot, few-shot, and chain-of-thought prompting. Although decompositional CoT markedly improves \textsf{Llama-3.1-8B} and \textsf{GPT-3.5-turbo} over their vanilla and $3$-shot counterparts, a pronounced gap to \textsf{AR-GCAE} persists on both datasets. The main obstacle is that these models consume the causality graph as a flattened token sequence and therefore struggle to perceive the multi-hop evolutionary dependencies among historical events. In contrast, \textsf{AR-GCAE} preserves graph structure while injecting retrieved abstract rules, letting it combine faithful topological reasoning with the broad causal priors that LLMs could only implicitly approximate, thereby converting the strengths of both relation-level and instance-level causal knowledge into a decisive advantage on the CGEP task.

\par
\textbf{Computational Efficiency.}
Besides prediction accuracy, we compare the computational cost of \textsf{AR-GCAE} against the representative encoder-based baselines, \textsf{SeDGPL} and \textsf{SEDA}, all measured under an identical hardware configuration on MAVEN-CGEP. As reported in Table~\ref{tab:compu}, despite injecting retrieved abstract rules, \textsf{AR-GCAE} introduces not much additional computational burden: with a comparable parameter budget ($382$M), it trains in $0.9$ hours per epoch within $10$GB of GPU memory, and its inference is the fastest among all methods ($8$ min per epoch). This indicates that the substantial accuracy gains delivered by AECRs come at essentially no extra computational cost, keeping \textsf{AR-GCAE} as efficient as, or more efficient than, existing baselines.

\begin{table}[t]
	\resizebox{\linewidth}{!}{
		\renewcommand{\arraystretch}{1}
		\begin{tabular}{@{}l|c|cc|cc}
			\toprule
			\midrule & & \multicolumn{2}{c}{\cellcolor[HTML]{f7f9f1}\textbf{Training phase}} & \multicolumn{2}{c}{\cellcolor[HTML]{f7f9f1}\textbf{Reasoning phase}} \\
			\cmidrule{3-6}
			\multirow{-2}{*}{\textbf{Model}} & \multirow{-2}{*}{\textbf{Param}} &
			\cellcolor[HTML]{fef8e6}Time/Epoch & \cellcolor[HTML]{fef8e6}GPU/Batch & \cellcolor[HTML]{fef8e6}Time/Epoch & \cellcolor[HTML]{fef8e6}GPU/Epoch \\ \midrule
			\textsf{SeDGPL} & 376M & 1.2hours & 12GB & 20min & 12GB \\
			\textsf{SEDA} & 255M & 0.5hours & 10GB & 10min & 10GB \\
            \textsf{TRACE} & 259M & 1.1hours & 19GB & 13min & 19GB \\
			\textsf{AR-GCAE} & 382M & 0.9hours & 10GB & 8min & 10GB \\
			\bottomrule
	\end{tabular}}
	\caption{Comparison of computational efficiency between AR-GCAE and strong baselines on the MAVEN-CGEP dataset.}
	\label{tab:compu}
    \vspace{-10pt}
\end{table}


\begin{figure}[tbp]  
		\centering
		\includegraphics[width=\columnwidth]{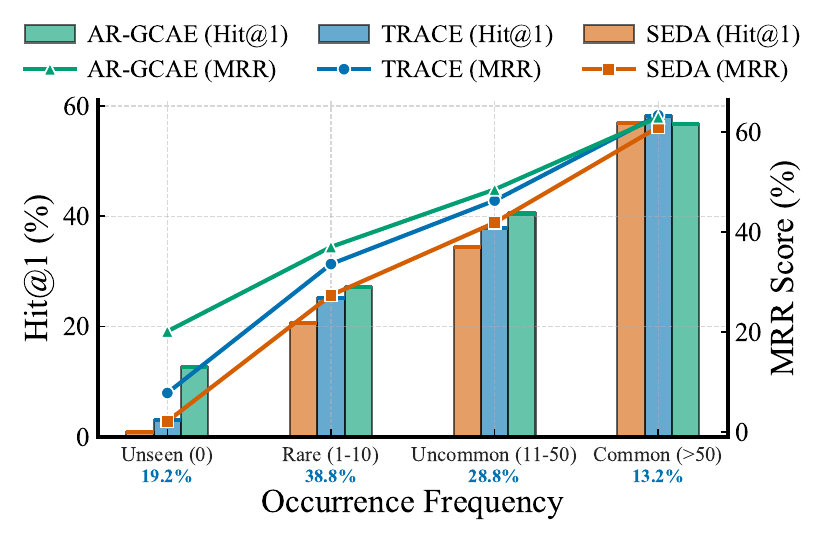}
		\caption{Results of frequency-stratified evaluation on MAVEN-CGEP: Hit@1 (grouped bars) and MRR (lines)}
		\label{Fig:longtail}  
\end{figure}

\subsection{Frequency-Stratified Evaluation}
  
\textbf{Experiments.} To investigate model performance under varying target event rarity, we partition MAVEN-CGEP test samples by the training-set frequency of their ground-truth consequent events into four non-overlapping groups: \textit{Common} (over $50$ training occurrences), \textit{Uncommon} ($11$–$50$), \textit{Rare} ($1$–$10$), and \textit{Unseen} ($0$, meaning the event never appears in training data). 

\par
\textbf{Results.}
Fig.~\ref{Fig:longtail} plots Hit@1 scores (grouped bars, left y-axis) and MRR values (line curves, right y-axis) for all models across each group, comparing our \textsf{AR-GCAE} against two strong baselines, \textsf{TRACE} and \textsf{SEDA}. The fraction of test samples within each group is annotated beneath each bar cluster. Crucially, low-frequency groups dominate the test distribution: Rare, Uncommon, and Unseen samples jointly account for $86.8\%$ of all test instances ($38.8\%$, $28.8\%$, and $19.2\%$, respectively), while the Common group occupies merely the remaining $13.2\%$. All methods yield comparable results on Common events. As event frequency falls into the Uncommon and Rare ranges, \textsf{TRACE} and \textsf{SEDA} suffer substantial performance degradation, whereas \textsf{AR-GCAE} undergoes gentle drops and steadily widens its performance lead. On the Unseen subset, the two baselines degrade to near-zero scores, yet \textsf{AR-GCAE} retains a prominent, consistent performance advantage.


\textbf{Analysis.}
These observations reveal that mainstream baselines merely memorize training data statistics and lack the capacity to generalize to out-of-distribution events. When target events appear sparsely or are entirely absent from training corpora, the instance-wise co-occurrence signals leveraged by baselines disappear completely, leading to near-zero performance. In contrast, \textsf{AR-GCAE} maintains a persistent performance advantage specifically on the Rare and Unseen groups, which validates the efficacy of our abstract causal rule base. Rather than relying on superficial instance-level pattern matching, our framework aligns novel event pairs with generalized causal rules. This abstract causal prior empowers precise logical extrapolation and equips the model with superior generalization ability toward long-tail and unseen cases.

	\begin{table}[t]
		\resizebox{\linewidth}{!}{
			\renewcommand{\arraystretch}{1}
			\begin{tabular}{@{}l|ccc|ccc}
				\toprule
				\midrule & \multicolumn{3}{c|}{\cellcolor[HTML]{f7f9f1}\textbf{MAVEN-CGEP}} & \multicolumn{3}{c}{\cellcolor[HTML]{f7f9f1}\textbf{ESC-CGEP}} \\ \cmidrule(lr){2-4} \cmidrule(lr){5-7}
				\multirow{-2}{*}{\textbf{Rule Source}} &
				\cellcolor[HTML]{fef8e6}MRR & \cellcolor[HTML]{fef8e6}Hit@1 & \cellcolor[HTML]{fef8e6}Hit@10 & \cellcolor[HTML]{fef8e6}MRR & \cellcolor[HTML]{fef8e6}Hit@1 & \cellcolor[HTML]{fef8e6}Hit@10 \\ \midrule
				\textsf{w/o AECR} & 34.8 & 27.1 & 51.5 & 19.3 & 15.2 & 27.5 \\
				\textsf{Cross-Dataset AECR} & \cellcolor[HTML]{f0f7ff}\underline{38.3} & \cellcolor[HTML]{f0f7ff}\underline{30.7} & \cellcolor[HTML]{f0f7ff}\underline{54.0} & \cellcolor[HTML]{f0f7ff}\underline{21.7} & \cellcolor[HTML]{f0f7ff}\underline{16.3} & \cellcolor[HTML]{f0f7ff}\underline{31.9} \\
				\midrule\midrule
				\textsf{In-Dataset AECR} & \cellcolor[HTML]{e0f0ff}\textbf{40.5} & \cellcolor[HTML]{e0f0ff}\textbf{32.1} & \cellcolor[HTML]{e0f0ff}\textbf{57.4} & \cellcolor[HTML]{e0f0ff}\textbf{22.8} & \cellcolor[HTML]{e0f0ff}\textbf{17.3} & \cellcolor[HTML]{e0f0ff}\textbf{32.4} \\ \midrule
				\bottomrule
		\end{tabular}}
		\caption{Results on cross-dataset rule transferability. }
		\label{tab:transfer}
        \vspace{-10pt}
	\end{table}

\subsection{Cross-Dataset Rule Transfer Evaluation}
\textbf{Experiments.}
To verify whether the efficacy of the AECR knowledge base is restricted to its source dataset, we conduct cross-dataset transfer experiments. Specifically, we perform event prediction on ESC-CGEP guided by the MAVEN-AECR rule base and, vice versa, run MAVEN-CGEP prediction using the ESC-AECR rule base. We name this cross-dataset configuration \textsf{Cross-Dataset AECR} and compare it with \textsf{In-Dataset AECR}, which adopts rules extracted from the matching dataset. We take the rule-free baseline \textsf{w/o AECR} (discussed in our ablation study) as a reference.

\textbf{Results \& Analysis.} Table~\ref{tab:transfer} reveals that \textsf{Cross-Dataset AECR} achieves steady and notable improvements over the baseline on both datasets. It boosts MRR from 34.8 to 38.3 on MAVEN-CGEP and from 19.3 to 21.7 on ESC-CGEP, recouping most performance gains delivered by \textsf{In-Dataset AECR}. These observations validate that causal knowledge stored in AECR knowledge bases is not merely dataset-specific artifacts, but transferable causal priors. Each rule abstracts concrete cause-effect pairs into relation-level causal schemas linking conceptual events, capturing invariant causal regularities that generalize across corpora with divergent surface events and separate domains. A causal rule abstracts concrete cause-effect pairs into relation-level schemas between conceptual events, capturing causal regularities that hold across corpora with distinct surface events and disparate domains. Accordingly, rules distilled from one dataset can deliver reliable causal guidance when applied to unseen target datasets. This further corroborates our core contribution that AECR yields generalizable, reusable causal knowledge rather than dataset-specific patterns.

    \begin{figure}[t]
    \centering
    \begin{subfigure}{\columnwidth}
        \centering
        \includegraphics[width=\columnwidth]{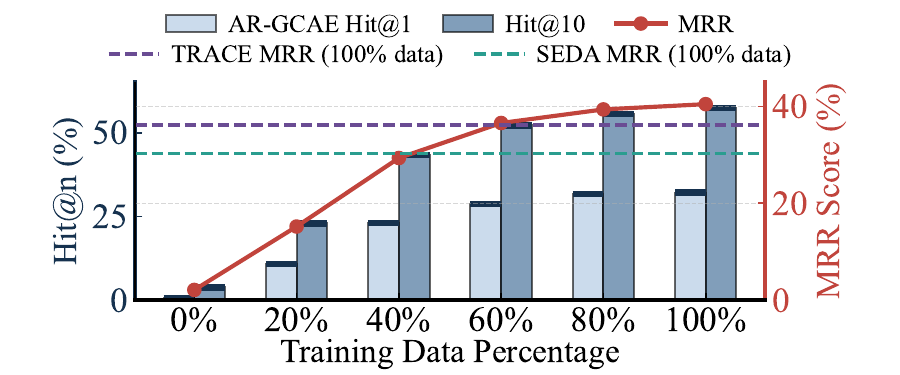}
        \caption{MAVEN-CGEP}
        \label{Fig:lowResource_maven}
    \end{subfigure}
    \\[4pt]
    \begin{subfigure}{\columnwidth}
        \centering
        \includegraphics[width=\columnwidth]{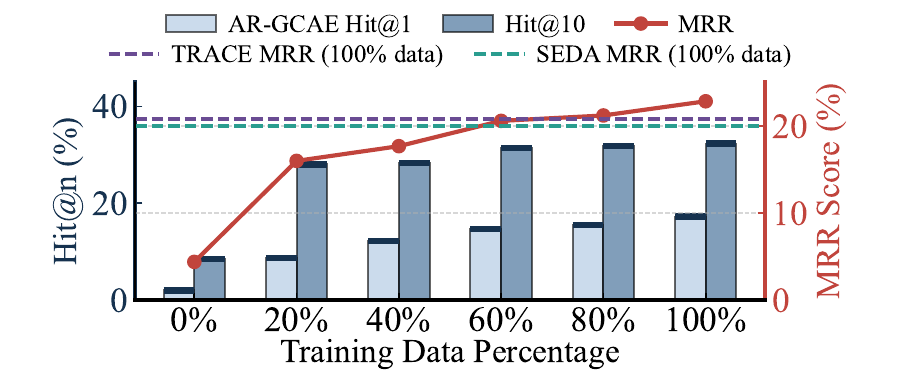}
        \caption{ESC-CGEP}
        \label{Fig:lowResource_esc}
    \end{subfigure}
    \caption{Results on low-resource scenario evaluation:  Hit@1 (grouped bars) and MRR (lines).}
    \label{Fig:lowResource}
    \vspace{-10pt}
\end{figure}


\begin{figure}[t]  
		\centering
		\includegraphics[width=\columnwidth]{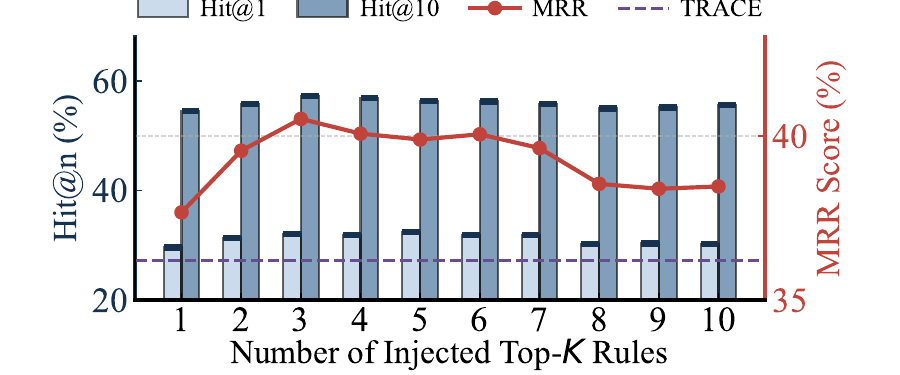}
		\caption{Results on the number of injected Top-$K$ rules on the MAVEN-CGEP dataset. }
		\label{Fig:K_abla}  
\end{figure}


\subsection{Low-Resource Scenario Evaluation}
\textbf{Results.}
As shown in Fig.~\ref{Fig:lowResource}, we vary the fraction of accessible training data from $0\%$ to $100\%$ and compare \textsf{AR-GCAE} against fully trained baselines. On both datasets, \textsf{AR-GCAE} improves steadily with more data and, using only a fraction of it, already matches or surpasses fully trained strong baselines: it exceeds \textsf{TRACE} ($36.2$) with $60\%$ of the data on MAVEN-CGEP (MRR $36.6$), and surpasses \textsf{TRACE} ($20.8$) and \textsf{SEDA} ($20.0$) with $80\%$ of the data on ESC-CGEP (MRR $21.2$).

\par
\textbf{Analysis.}
This observation verifies that our abstract causal rules substantially strengthen model robustness under low-resource constraints. The retrieved AECRs introduce transferable relation-level causal schemas, and accordingly, our \textsf{AR-GCAE} mitigates reliance on abundant task-specific labeled data and maintains competitive prediction performance even under data scarcity.

\begin{table}[t]
		\resizebox{\linewidth}{!}{
			\renewcommand{\arraystretch}{1}
			\begin{tabular}{@{}l|ccc|ccc}
				\toprule
				\midrule & \multicolumn{3}{c|}{\cellcolor[HTML]{f7f9f1}\textbf{MAVEN-CGEP}} & \multicolumn{3}{c}{\cellcolor[HTML]{f7f9f1}\textbf{ESC-CGEP}} \\ \cmidrule(lr){2-4} \cmidrule(lr){5-7}
				\multirow{-2}{*}{\textbf{Model}} &
				\cellcolor[HTML]{fef8e6}MRR & \cellcolor[HTML]{fef8e6}Hit@1 & \cellcolor[HTML]{fef8e6}Hit@10 & \cellcolor[HTML]{fef8e6}MRR & \cellcolor[HTML]{fef8e6}Hit@1 & \cellcolor[HTML]{fef8e6}Hit@10 \\ \midrule
				\textsf{w/o AECR} & 34.8 & 27.1 & 51.5 & 19.3 & 15.2 & 27.5 \\
				\textsf{w/o Gated Fusion} & \cellcolor[HTML]{f0f7ff}\underline{38.8} & \cellcolor[HTML]{f0f7ff}\underline{30.3} & \cellcolor[HTML]{f0f7ff}\underline{56.4} & \cellcolor[HTML]{f0f7ff}\underline{21.9} & \cellcolor[HTML]{f0f7ff}\underline{16.8} & \cellcolor[HTML]{f0f7ff}\underline{31.3} \\
				\textsf{w Random Rules} & 35.0 & 27.1 & 51.6 & 19.7 & 15.0 & 28.5 \\
				\midrule\midrule
				\textsf{Full \textbf{AR-GCAE}} &
                \cellcolor[HTML]{e0f0ff}\textbf{40.5} & \cellcolor[HTML]{e0f0ff}\textbf{32.1} & \cellcolor[HTML]{e0f0ff}\textbf{57.4} & \cellcolor[HTML]{e0f0ff}\textbf{22.8} & \cellcolor[HTML]{e0f0ff}\textbf{17.3} & \cellcolor[HTML]{e0f0ff}\textbf{32.4} \\ \midrule
				\bottomrule
		\end{tabular}}
		\caption{Ablation study results on two datasets.}
		\label{tab:ablation}
\end{table}
\subsection{Ablation Study}
\textbf{Results.}
Table~\ref{tab:ablation} provides the overall performance of the full framework and its three variants. The complete model achieves the best results across all metrics. Removing the abstract rules (\textsf{w/o AECR}) causes the largest degradation. Discarding the gated fusion module (\textsf{w/o Gated Fusion}), so that predictions rely solely on the rule-guided re-encoding, leads to a moderate but consistent decline. Injecting randomly sampled rules instead of retrieved ones (\textsf{w Random Rules}) yields a performance level nearly identical to that of entirely removing the causal rule.

\par
\textbf{Analysis.}
These results reveal the distinct role of each component. First, the sharp drop of the \textsf{w/o AECR} variant confirms that the abstract causal rules provide indispensable relation-level knowledge beyond what graph topology alone can encode. Second, the decline of the \textsf{w/o Gated Fusion} variant indicates that accurate prediction requires both the rule-free representation and the rule-guided re-encoding: relying on the rule-guided branch alone discards complementary evidence carried by the original encoding. Third, and most notably, injecting random causal rules (\textsf{w Random Rules}) does not degrade performance below the \textsf{w/o AECR} level, showing that spurious rules do not mislead the prediction network. We attribute this robustness to the gated fusion mechanism, which adaptively balances how much it trusts the rule-free and rule-guided representations, and thus suppresses the influence of noisy or irrelevant rules. Together, these findings demonstrate that the AECR knowledge base supplies the essential causal knowledge for generalization, while the gated fusion module governs how this knowledge is integrated.

\subsection{Impact of The Number of Retrieved Rules}
\textbf{Results.}
Fig.~\ref{Fig:K_abla} reports how AR-GCAE behaves on MAVEN-CGEP as the number of injected Top-$K$ rules varies from $1$ to $10$, with MRR drawn as a line against the right axis and Hit@1 and Hit@10 as clustered bars against the left axis. As $K$ grows, all three metrics first rise rapidly and then decline gently: starting from $K=1$, MRR climbs to its peak of $40.5$ at $K=3$ (where Hit@10 peaks as well), after which the metrics fluctuate slightly and drift downward as $K$ continues to increase. Crucially, \textsf{AR-GCAE} stays clearly ahead of the strongest baseline \textsf{TRACE} (dashed line) across the entire range of $K$: even its weakest configuration ($K=1$, MRR $37.7$) already surpasses \textsf{TRACE} (MRR $36.2$).

\par
\textbf{Analysis.}
This trend reflects a trade-off between information gain and semantic noise. When $K$ is small, enlarging the retrieved set raises the probability of covering the correct causal rule, which supplies essential causal knowledge and drives the rapid initial improvement. Beyond the optimum, however, each additional rule increasingly introduces irrelevant content that dilutes the attention over genuinely relevant rules, causing the mild subsequent decline. Notably, this degradation is gradual rather than catastrophic, and \textsf{AR-GCAE} remains ahead of \textsf{TRACE} regardless of the choice of $K$, indicating that our method is robust to this hyperparameter.


%% file: Sections/Conclusion.tex
\section{Conclusion}\label{Sec:Conclusion}
This paper presents the \textit{Abstract Event Causal Rule} (AECR), a relation-level abstraction that converts concrete cause-effect event pairs into transferable causal schemas while retaining their inherent causal connections. We distill two AECR knowledge bases via our multi-agent CACI framework; human evaluation validates the rationality, discriminability and practical utility of the constructed knowledge bases. Further, we integrate retrieved AECR rules into event prediction tasks via our proposed \textsf{AR-GCAE} encoder. Comprehensive experimental comparisons against state-of-the-art methods yield consistent performance improvements. Our approach exhibits distinct advantages on long-tail and unseen events, and it also delivers strong cross-dataset transferability and favorable robustness under low-resource constraints.

\par
Our future work aims to expand the applicability of AECR to zero-shot causal reasoning in more domains such as finance and risk management. Further, we plan to build an early warning system that operates without task-specific retraining by aligning domain-specific event instances with our universal abstract causal rules.